\documentclass[sigconf]{acmart}
\setcitestyle{numbers,sort&compress}

\title{ BenGER Platform: A Collaborative Web Platform for End-to-End Benchmarking of German Legal Tasks}

\author{Sebastian Nagl}
\affiliation{
  \institution{Technical University of Munich}
  \city{Munich}
  \country{Germany}
}
\email{sebastian.nagl@tum.de}

\author{Matthias Grabmair}
\affiliation{
  \institution{Technical University of Munich}
  \city{Munich}
  \country{Germany}
}
\email{matthias.grabmair@tum.de}

\acmConference[ICAIL 2026]{International Conference on Artificial Intelligence and Law}{June 08-12}{Singapore}
\acmYear{2026}
\copyrightyear{2026}

\acmDOI{}
\acmISBN{}
\acmSubmissionID{}
\setcopyright{none}
\renewcommand\footnotetextcopyrightpermission[1]{}
\begin{abstract}
Evaluating large language models (LLMs) for legal reasoning requires workflows that span task design, expert annotation, model execution, and metric-based evaluation. In practice, these steps are split across platforms and scripts, limiting transparency, reproducibility, and participation by non-technical legal experts.
We present the \textbf{BenGER}\footnote{The full code will be released as free software on acceptance. A public instance of the application is accessible at \url{https://what-a-benger.net}.} (Benchmark for German Law) framework, an open-source web platform that integrates task creation, collaborative annotation, configurable LLM runs, and evaluation with lexical, semantic, factual, and judge-based metrics. BenGER supports multi-organization projects with tenant isolation and role-based access control, and can optionally provide formative, reference-grounded feedback to annotators. We will demonstrate a live deployment showing end-to-end benchmark creation and analysis.
\end{abstract}


\begin{document}
\maketitle

\section{Introduction and Motivation}
Legal AI benchmarking is costly and technically demanding, especially in jurisdictions such as Germany where high-quality legal expertise is scarce. In many projects, the pipeline is fragmented: experts provide materials or it is taken from existing sources, researchers translate them into benchmark instances, optionally annotations are collected in separate tooling, model runs are executed via ad-hoc scripts, and evaluation code is reimplemented per study \citep{fan2025, guha2022}. This introduces avoidable handoffs, reduces expert oversight, and makes collaboration and reproduction difficult.
BenGER addresses this by providing a unified, browser-based workflow that domain experts can operate end-to-end, from task definition and annotation through model execution and evaluation.

\section{System Overview}
BenGER is a production-ready web application for end-to-end benchmarking of legal tasks (initially focused on German law, but not jurisdiction-bound). It supports multiple task formats (free-text reasoning, multiple choice, span annotation), collaborative annotation, batch execution of arbitrary LLMs, and result analysis using a broad set of metrics. BenGER is released as open-source software and can be deployed locally or institutionally while we also offer a securely hosted version for our community work.

\section{Supported Workflow and Use Cases}

BenGER explicitly models the full legal benchmarking workflow that we will also show in our demonstration:

\begin{enumerate}
  \item \textbf{Task Creation}: Legal experts define tasks and reference solutions directly in the platform.
  \item \textbf{Annotation}: Human annotators submit solutions using a collaborative web interface.
  \item \textbf{Formative Feedback (Optional)}: Annotators may receive LLM-based feedback comparing their answers to reference solutions, providing constructive guidance.
  \item \textbf{Model Execution}: Selected LLMs are executed on the same tasks using configurable API keys.
  \item \textbf{Evaluation}: Results are evaluated using lexical, semantic, factual, classification, and LLM-as-a-judge metrics.
  \item \textbf{Analysis and Export}: Results can be analyzed within the platform or exported for publication.
\end{enumerate}

This workflow supports collaborative research scenarios involving universities, public authorities, NGOs, and individual legal experts.

\section{Technical Architecture}
BenGER uses a modular service architecture with a Next.js (TypeScript) frontend and a FastAPI (Python) backend backed by PostgreSQL. 
Redis and Celery workers support scalable background execution for model runs and evaluations. The system is fully containerized and deployable via Docker Compose or Kubernetes.

\section{Data Governance, Security, and Responsible Use}
BenGER is designed for collaborations involving potentially sensitive legal materials. It enforces tenant isolation and role-based permissions, enabling fine-grained sharing without cross-organization leakage. Model execution can be configured per user or per project, including support for user-provided API credentials to align with institutional policies. Optional LLM feedback is intended as supportive, educational guidance and can be disabled or restricted per project.

\paragraph{Tenant isolation and access control.}
Organizations are isolated at the data layer and via role-based permissions (e.g., administrators, contributors, annotators). Project-level access controls allow fine-grained sharing while preventing accidental cross-organization data exposure.

\paragraph{API key handling and operational boundaries.}
Model execution can be configured per user or per project, enabling contributors to use their own API credentials when appropriate. This reduces centralized credential management and helps align usage with institutional policies.

\paragraph{Human oversight.}
Optional LLM feedback is designed to be supportive and educational rather than authoritative. Projects can disable feedback entirely or restrict it to specific tasks and roles, ensuring that expert governance remains the primary mechanism for benchmark quality assurance.

\section{Positioning and Benefits over Existing Tooling}
BenGER targets a common gap: existing solutions often cover single steps of the benchmarking pipeline - \emph{either} annotation/data management (e.g. DeepWrite \citep{kramerDeepWrite2024}) \emph{or} model evaluation, but do not provide an integrated, role-aware workflow that domain experts can run without scripting.
Compared to general-purpose annotation platforms (such as LabelStudio or Doccano \citep{labelstudio,doccano}), BenGER adds multi-organization isolation, configurable LLM execution, and standardized evaluation runs within one system and cost-free. Compared to ad-hoc evaluation scripts, it turns tasks, model configurations, and metrics into reusable, auditable artifacts, improving reproducibility for groups spanning universities, public institutions, and NGOs.

\paragraph{Beyond general-purpose annotation platforms.}
General legal annotation systems (e.g. Lawnotation \citep{vandijkLawnotation2022}) offer flexible labeling UIs and dataset export, but they generally require additional infrastructure to (a) enforce clean separation between multiple contributing organizations, (b) connect to heterogeneous LLM providers, and (c) execute and track evaluation runs with pre-defined and therefore consistent metric definitions.
BenGER closes this gap by combining native legal-task annotation with model execution and standardized evaluation, allowing domain experts to retain control over task definition, reference answers, and quality assurance throughout the lifecycle.

\paragraph{Beyond evaluation scripts and ad-hoc pipelines.}
In many research projects, evaluation is implemented as project-specific code: prompt templates, model calls, and metrics are encoded in notebooks or scripts that are hard to reuse across organizations and tasks.
BenGER externalizes these steps into a shared platform: tasks, model configurations, and metrics become explicit artifacts that can be reused, audited, and compared across groups, improving reproducibility and lowering onboarding costs.

\section{Benefits for Annotators and Human Baselines}
BenGER supports reliable human baselines with progress tracking and quality signals (e.g., agreement/consistency indicators). To improve incentives and learning value, the platform can optionally provide reference-grounded, constructive feedback to annotators - typical for german legal education (from a private lecturer, the 'Repetitor'), highlighting missing reasoning steps and common pitfalls while keeping expert governance in control.

\paragraph{Human baselines with quality signals.}
The platform supports quality monitoring at the annotation level (e.g., progress tracking and agreement/consistency indicators), enabling project leads to manage baseline construction systematically. In practice, this reduces the risk that benchmark conclusions are driven by noisy annotations or inconsistent task interpretation.

\section{Community Impact and Reproducibility}
As open-source software, BenGER lowers barriers for non-technical contributors by making task contribution, annotation, model execution, and evaluation accessible via a browser. It promotes reproducibility by storing task definitions, reference solutions, model configurations, and metric choices as explicit artifacts that can be shared, audited, and rerun. The integration layers are designed to be extensible for new tasks, model providers, and metrics.

\paragraph{Lowering barriers for public institutions and NGOs.}
Legal datasets are often constrained by capacity and governance requirements. By providing organization-aware data separation and a browser-based workflow for task creation, annotation, and evaluation, BenGER enables institutions to contribute tasks and obtain model performance analyses without handing off raw materials to external engineers.

\paragraph{Reproducible evaluation artifacts.}
BenGER encourages evaluation configurations to be stored as explicit, shareable artifacts: task definitions, reference solutions, model configurations, and chosen metrics. This supports transparent reporting and makes it easier to reproduce experimental results across research groups and over time.

\paragraph{Extensibility.}
The metric and model integration layers are designed for incremental extension. New tasks, model providers, or scoring methods can be added without rewriting end-to-end evaluation pipelines, making the platform suitable for long-lived benchmark initiatives.

\section{Conclusion}

BenGER demonstrates how legal AI benchmarking can be made more transparent, collaborative, and accessible by integrating annotation, model evaluation, and analysis into a single platform. We make legal experts control the full evaluation pipeline, which makes the system supports more reliable and can lead to scalable research on LLM capabilities in law.

\bibliographystyle{ACM-Reference-Format}
\bibliography{references}

@misc{LabelStudio,
  title={{Label Studio}: Data labeling software},
  url={https://github.com/HumanSignal/label-studio},
  note={Open source software available from https://github.com/HumanSignal/label-studio},
  author={
    Maxim Tkachenko and
    Mikhail Malyuk and
    Andrey Holmanyuk and
    Nikolai Liubimov},
  year={2020-2025},
}

@misc{doccano,
  title={{doccano}: Text Annotation Tool for Human},
  url={https://github.com/doccano/doccano},
  note={Software available from https://github.com/doccano/doccano},
  author={
    Hiroki Nakayama and
    Takahiro Kubo and
    Junya Kamura and
    Yasufumi Taniguchi and
    Xu Liang},
  year={2018},
}

@misc{vandijkLawnotation2022,
  author       = {van Dijck, Gijs and Aguilera, Carlos and van der Lans, Chris and Chakravarthy, Shashank and van Essel, Sander},
  title        = {Lawnotation: A Formal Language for Legal Rules},
  year         = {2022},
  howpublished = {\url{https://www.lawnotation.org/}}
}

@misc{kramerDeepWrite2024,
  author       = {Kramer, Urs and Granitzer, Michael and Graf Lambsdorff, Johann},
  title        = {DeepWrite: Annotation and Extraction of Legal Texts},
  year         = {2024},
  howpublished = {\url{https://extract-annotations.deepwrite.pads.fim.uni-passau.de/}},
  institution  = {University of Passau}
}

@misc{fan2025,
  title = {{{LEXam}}: {{Benchmarking Legal Reasoning}} on 340 {{Law Exams}}},
  shorttitle = {{{LEXam}}},
  author = {Fan, Yu and Ni, Jingwei and Merane, Jakob and Salimbeni, Etienne and Tian, Yang and Hermstr{\"u}wer, Yoan and Huang, Yinya and Akhtar, Mubashara and Geering, Florian and Dreyer, Oliver and Brunner, Daniel and Leippold, Markus and Sachan, Mrinmaya and Stremitzer, Alexander and Engel, Christoph and Ash, Elliott and Niklaus, Joel},
  year = 2025,
  month = may,
  number = {arXiv:2505.12864},
  eprint = {2505.12864},
  primaryclass = {cs},
  publisher = {arXiv},
  doi = {10.48550/arXiv.2505.12864},
  urldate = {2025-06-01},
  archiveprefix = {arXiv},
  langid = {english}
}

@misc{guha2022,
  title = {{{LegalBench}}: {{Prototyping}} a {{Collaborative Benchmark}} for {{Legal Reasoning}}},
  shorttitle = {{{LegalBench}}},
  author = {Guha, Neel and Ho, Daniel E. and Nyarko, Julian and R{\'e}, Christopher},
  year = 2022,
  month = sep,
  number = {arXiv:2209.06120},
  eprint = {2209.06120},
  primaryclass = {cs},
  publisher = {arXiv},
  urldate = {2023-08-02},
  archiveprefix = {arXiv}
}

\end{document}